\documentclass[10pt,twocolumn,letterpaper]{article}

\usepackage{wacv}
\usepackage{times}
\usepackage{epsfig}
\usepackage{graphicx}
\usepackage{amsmath}
\usepackage{amssymb}

\usepackage[utf8]{inputenc}
\usepackage[percent]{overpic}
\usepackage[table]{xcolor}
\usepackage[export]{adjustbox}
\usepackage{multirow}
\usepackage{enumitem}
\usepackage{array}
\usepackage{dblfloatfix}

\makeatletter
\@namedef{ver@everyshi.sty}{}
\makeatother
\usepackage{tikz}
\usepackage{pgfplots}

\makeatletter
\let\MYcaption\@makecaption
\makeatother

\usepackage[font=footnotesize]{subcaption}

\makeatletter
\let\@makecaption\MYcaption
\makeatother

\definecolor{burntorange}{rgb}{0.8,0.33,0.0}
\definecolor{skyblue}{rgb}{0.53,0.81,0.98}

\newcommand\Tstrut{\rule{0pt}{2.6ex}}         
\newcommand\Bstrut{\rule[-0.8ex]{0pt}{0pt}}   

\newcolumntype{R}[2]{%
    >{\adjustbox{angle=#1,lap=\width-(#2)}\bgroup}%
    c%
    <{\egroup}%
}
\newcommand*\rot{\multicolumn{1}{R{25}{1em}}}
\newcolumntype{C}{>{\centering\let\newline\\\arraybackslash\hspace{0pt}}m{2.25em}}

\newcommand{\name}{SSGP}

\def\wacvPaperID{42} 

\wacvfinalcopy 

\ifwacvfinal
\def\assignedStartPage{1} 
\fi

\usepackage[pagebackref=true,breaklinks=true,colorlinks,bookmarks=false]{hyperref}

\usepackage[sort&compress,capitalize,noabbrev,nameinlink]{cleveref}
\creflabelformat{equation}{#2#1#3}

\ifwacvfinal
\else
\fi

\begin{document}

\title{SSGP: Sparse Spatial Guided Propagation\\for Robust and Generic Interpolation}

\author{René Schuster\textsuperscript{1} \hspace{0.5cm} Oliver Wasenmüller\textsuperscript{1} \hspace{0.5cm} Christian Unger\textsuperscript{2} \hspace{0.5cm} Didier Stricker\textsuperscript{1} \\
\textsuperscript{1}DFKI - German Research Center for Artificial Intelligence \hspace{5mm}
\textsuperscript{2}BMW Group \\
{\tt\small firstname.lastname@\string{dfki,bmw\string}.de}
}


\maketitle

\begin{abstract}
Interpolation of sparse pixel information towards a dense target resolution finds its application across multiple disciplines in computer vision. State-of-the-art interpolation of motion fields applies model-based interpolation that makes use of edge information extracted from the target image. For depth completion, data-driven learning approaches are widespread. Our work is inspired by latest trends in depth completion that tackle the problem of dense guidance for sparse information. We extend these ideas and create a generic cross-domain architecture that can be applied for a multitude of interpolation problems like optical flow, scene flow, or depth completion. In our experiments, we show that our proposed concept of Sparse Spatial Guided Propagation (SSGP) achieves improvements to robustness, accuracy, or speed compared to specialized algorithms.
\end{abstract}

\section{Introduction} \label{sec:intro}

The problems of interpolation and extrapolation have a long history in mathematics and computer science.
In high-level computer vision, interpolation finds its application in various problems like motion estimation in 2D (optical flow) \cite{bailer2015flow,bailer2019flow,gibson2003robust,hu2017robust,hu2016efficient,lang2012practical,nicolescu2003layered,oliver2018motion,revaud2015epic,wang2019semflow,zweig2017interponet}, 3D (scene flow) \cite{schuster2018sceneflowfields,schuster2020sffpp}, or depth completion \cite{cheng2018depth,jaritz2018sparse,ma2019self,tang2019learning,uhrig2017sparsity}.
These methods in turn are applied in robot navigation, advanced driver assistance systems (ADAS), surveillance, and many others.

The strategies of previous work are quite distinct for motion field interpolation and depth completion.
While the first focuses on hand-crafted models and piece-wise patches extracted from edge information, the latter fully relies on deep neural networks often considering image information insufficiently.
With the learning capabilities and inherent parallelism of the data-driven approach, we want to further push the limits of motion field estimation towards higher accuracy and speed.
At the same time, we extend and combine previous ideas from depth completion into a model that works equally well on different domains and applications.
This exposes novel challenges like effective mechanisms for handling of sparse data with different patterns or densities, efficient strategies for guidance from dense image information, or suitable fusion of heterogeneous data (\eg image and depth feature representations).

\begin{figure}[t]
	\centering
	\begin{subfigure}[c]{0.78\linewidth}
		\includegraphics[width=1\linewidth]{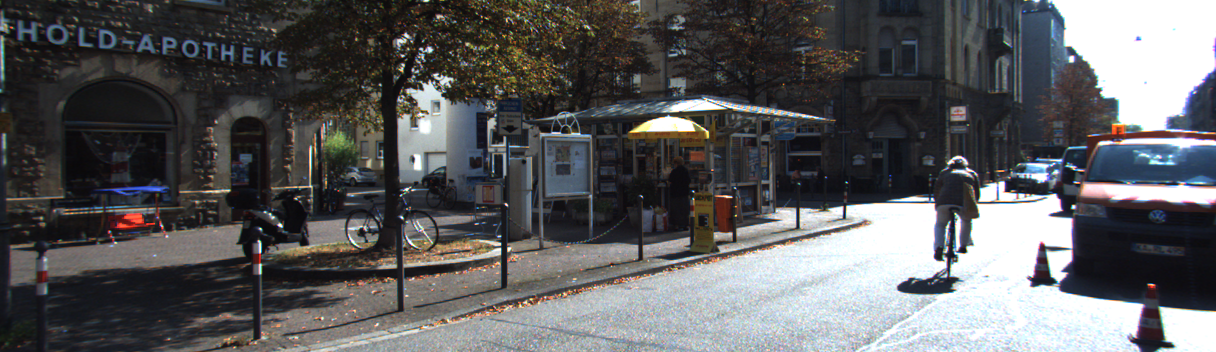}
		\caption{Input image}
		\vspace{1mm}%
	\end{subfigure}
	\begin{subfigure}[c]{0.78\linewidth}
		\includegraphics[width=1\linewidth]{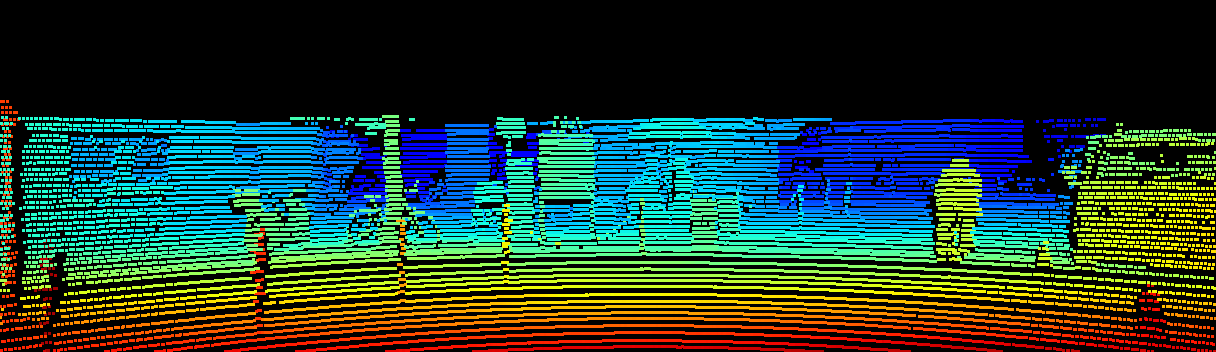}
		\caption{LiDAR measurements (visually enhanced)}
		\vspace{1mm}%
	\end{subfigure}\\%
	\begin{subfigure}[c]{0.78\linewidth}
		\includegraphics[width=1\linewidth]{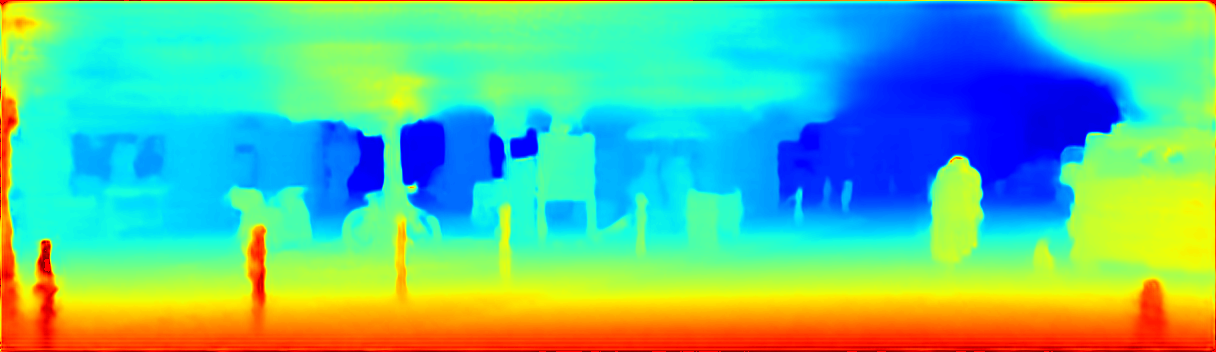}
		\caption{Densified depth with SSGP}
		\vspace{1mm}%
	\end{subfigure}
	\caption{We propose Sparse Spatial Guided Propagation (\name{}), a deep network for interpolation of sparse data. Here, an example of depth completion on KITTI \cite{geiger2012kitti} data is shown. Our full evaluation conducts experiments on more data sets and different types of input.}
	\label{fig:title}
	\vspace{-2mm}
\end{figure}

To solve the aforementioned challenges, we propose Sparse Spatial Guided Propagation (\name{}), which is the combination of spatially invariant, image dependent \textit{convolutional propagation} and \textit{sparsity-aware convolution}.
This key concept is used in a \textit{generic} sparse-to-dense encoder-decoder with \textit{full image guidance} at every stage.
Our overall contribution consists of the following:
\begin{itemize}[noitemsep,topsep=1pt,leftmargin=*,label={\tiny\raisebox{0.75ex}{\textbullet}}]
	\item A unified architecture which performs sparse-to-dense interpolation in different domains, \eg interpolation of optical flow, scene flow, or depth.
	\item A proper architectural design that leads to excellent robustness against noisy input or changes in the input density.
	\item Appropriate image guidance to resolve the dependency of previous flow interpolators on edge maps.
	\item A modification of existing spatial propagation that saves a vast amount of trainable parameters and improves generalization.
	\item Exhaustive experiments to validate all the above claims and to compare to state-of-the-art where in several cases \name{} produces top results.
\end{itemize}

\section{Related Work} \label{sec:relatedwork}

\paragraph{Sparse-to-Dense Motion Estimation.}
The interpolation of sparse points to a dense motion field dates back to at least \cite{gibson2003robust,nicolescu2003layered}.
A practical approach for large displacement optical flow is introduced by EPICFlow \cite{revaud2015epic}.
The authors make use of image edges computed with SED \cite{zbontar2015computing} to find local edge-aware neighborhoods of previously computed, sparse flow values.
Based on these neighborhoods, an affine 2D transformation is estimated to interpolate the gaps.
Later, this concept is improved by RICFlow \cite{hu2017robust} to be more robust by using small superpixels and RANSAC in the estimation of the transformation.
SFF \cite{schuster2018sceneflowfields} and SFF++ \cite{schuster2020sffpp} take both interpolators for optical flow and transfer them to the scene flow setup.
Throughout this work, we will refer to the interpolation modules of SFF and SFF++ as \textit{EPIC3D} and \textit{RIC3D} respectively.
SemFlow \cite{wang2019semflow} extends the above concepts for interpolation of optical flow by the use of deeply regressed semantic segmentation maps.
These maps replace the edge information used in EPIC or RIC to improve the measure of similarity of connected neighborhoods of input matches.
However, this approach is heavily dependent on semantic segmentation algorithms and thus not suitable for all domains and data sets.
Lastly, InterpoNet \cite{zweig2017interponet} is another recent approach that considers deep neural networks for the actual interpolation task.
Yet, InterpoNet still requires an explicit edge map as input.

In contrast to all interpolation modules mentioned, our network performs dense interpolation at full resolution for a multitude of problems (\ie it is not restricted to optical flow or scene flow) and utilizes a trainable deep model (\ie it is not subjected to hand-crafted rules or assumptions and provides significantly better run-times).
Additionally, the existing approaches highly depend on an intermediate representation of the image (edges, semantics).
\name{} operates on the input image directly and resolves this dependency.

\paragraph{Depth Completion.}
Most recent related work (especially in the area of deep learning) is concerned with depth completion.
In this field, literature differentiates between unguided and guided depth completion.
The latter utilizes the reference image for guidance.
In the setup of guided depth completion, novel questions arise which are also highly relevant for this work, \eg how to deal with sparse information in neural networks or how to combine heterogeneous feature domains.
SparseConvNet \cite{uhrig2017sparsity} introduces sparsity invariant CNNs by normalizing regular convolutions according to a sparsity mask.
This work has also introduced the Depth Completion Benchmark to the KITTI Vision Benchmark Suite \cite{geiger2012kitti}. 
Later, another strategy for the handling of sparsity was introduced by confidence convolution \cite{eldesokey2019confidence}.
In this case, the authors replace the binary sparsity mask with a continuous confidence volume that is used to normalize features after convolution.

Another promising strategy is the use of spatially variant and content dependent kernels in convolutional networks \cite{li2016deep,wu2018fast}.
This idea is successfully used by \cite{liu2017learning} for semantic segmentation and later by CSPN \cite{cheng2018depth} for the refinement of already densified depth maps.
Most recently, GuideNet \cite{tang2019learning} has applied the same idea for the densification of sparse depth maps itself.
In all cases, the idea is to predict per-pixel propagation kernels based on the image (or a feature map) directly instead of learning a spatially invariant set of kernels that is likewise applied to every pixel of the input.

We will make use of the two latterly presented concepts, namely awareness and explicit handling of sparsity as well as learning of spatially-variant and image-dependent convolutions. Both ideas will be combined in our novel, sparsity-aware, image-guided interpolation network that uses our new Sparse Spatial Guided Propagation (\name{}) module.

\paragraph{Other Interpolation Tasks.}
Lastly, there are more computer vision problems that are remotely related to our work, \eg image inpainting which is also a problem of interpolation.
However, for image inpainting the challenge usually lies within the reconstruction of the texture.
For the interpolation of geometry or motion, the expected result is piece-wise smooth and thus the problem is rather to find semantically coherent regions.
Still, related ideas can also be found in the field of image inpainting, where \eg in \cite{liu2018image} partial convolutions are used, which is the same idea for handling of sparsity as in \cite{uhrig2017sparsity}.
Similarly, the task of super-resolution could also be posed as an interpolation problem with a regular pattern of sparse input.
Though theoretically, our method is directly applicable to this family of problems, super-resolution goes beyond the scope of this paper and might be easier to be solved with other approaches.

\begin{figure*}[t]
	\centering
	\begin{subfigure}[c]{0.98\linewidth}
		\centering
		\includegraphics[scale=0.6]{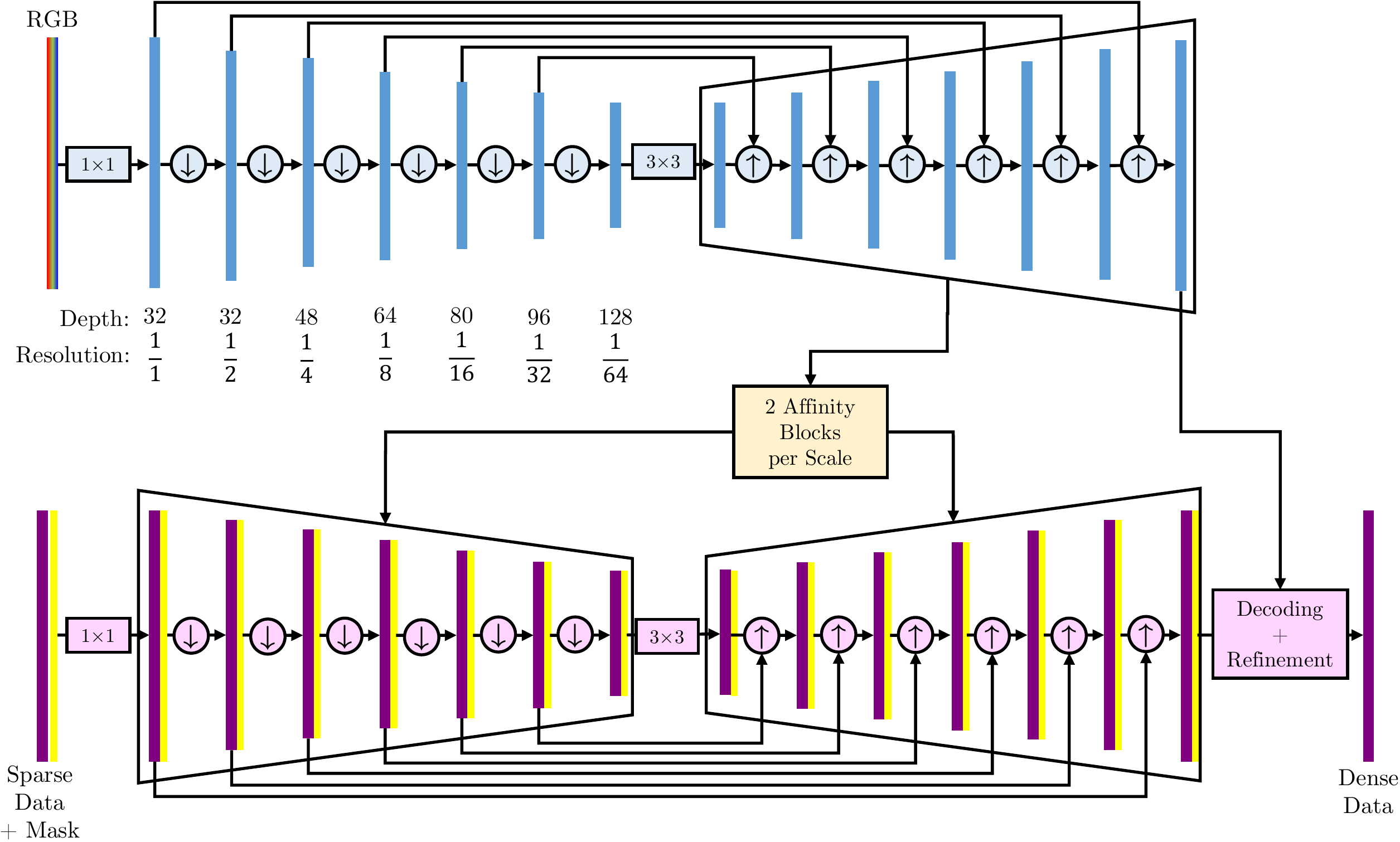}%
		\caption{Overall network architecture showing the RGB and the sparse-to-dense codec.}
		\label{fig:network:overview}
		\vspace*{6mm}
	\end{subfigure}
	\begin{subfigure}[t]{0.48\linewidth}
		\centering
		\includegraphics[scale=0.6]{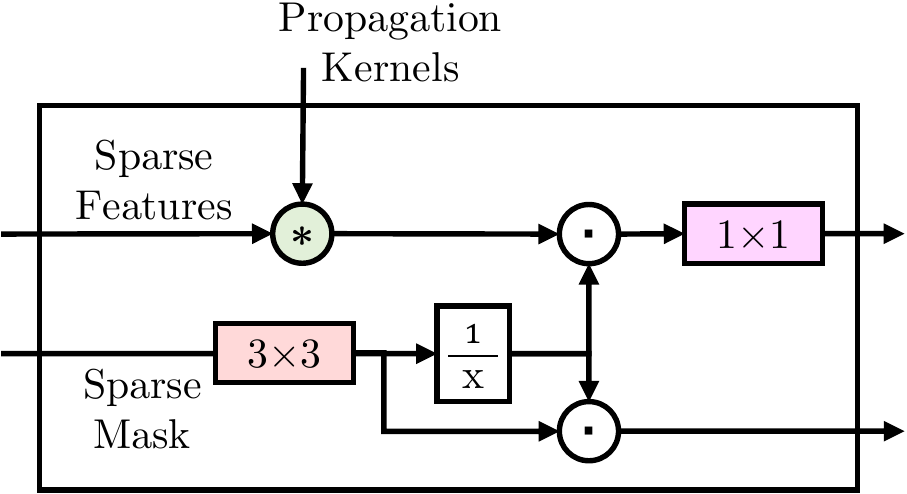}%
		\caption{Our novel sparse spatial propagation module.}
		\label{fig:network:sparsespatialpropagation}
		\vspace{3mm}%
	\end{subfigure}
	\hspace{0.02\linewidth}
	\begin{subfigure}[t]{0.48\linewidth}
		\centering
		\includegraphics[scale=0.6]{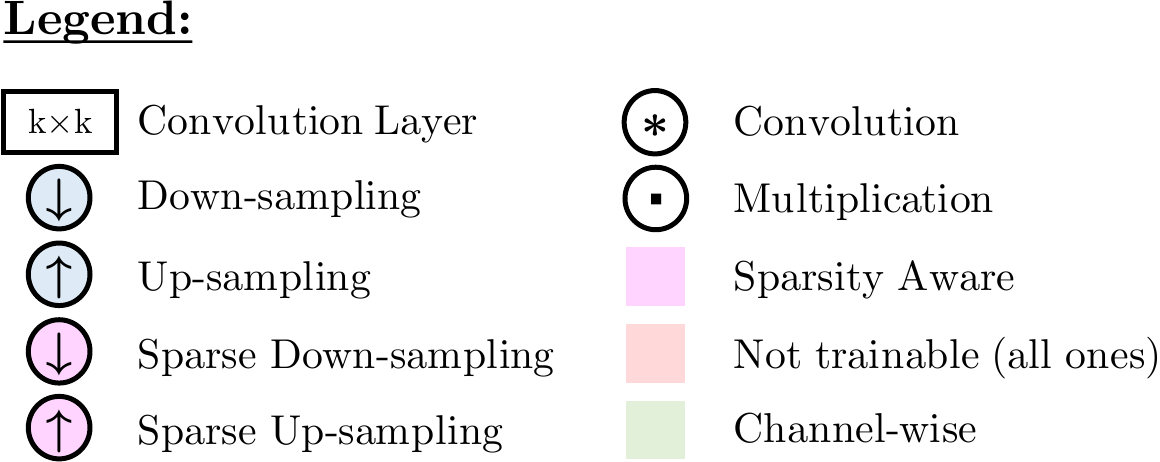}%
		\label{fig:network:legend}	
	\end{subfigure}
	\caption{An overview of our network architecture (\subref{fig:network:overview}) as well as a close-up view on our sparse spatial propagation module (\subref{fig:network:sparsespatialpropagation}) which is used in the down- and up-sampling blocks of the sparse-to-dense codec.}
	\label{fig:network}
\end{figure*}

\section{Interpolation Network} \label{sec:ssp}
As motivated earlier, we will use a deep neural network for the task of sparse-to-dense interpolation.
The network has to be equipped with an appropriate mechanism for sparsity, otherwise the considerably large gaps in the used sparse-to-dense motion estimation pipelines can lead to significantly deteriorated feature representation in these regions.
For the same reason of large gaps in motion fields (contrary to \eg depth completion where LiDAR measurements follow a predictable pattern of rotated scan lines), the network architecture has U-Net \cite{ronneberger2015unet} structure.
This way, even large gaps will be effectively closed after a few levels of the encoder, leading to a dense representation at the bottleneck.
Additionally, to inject a maximal amount of guidance through the entire sparse-to-dense codec,  the image information is used to compute spatially variant propagation kernels that are applied for densification by convolutional propagation in the sparse encoder, and for guided up-sampling in the dense decoder.
These guidance kernels are computed from the RGB image within a feature pyramid network with skip connections, for high expressiveness and accurate localization.

In summary, the interpolation network consists of four components.
Firstly, the RGB codec for computation of image-dependent and spatially-variant propagation kernels (\cref{sec:rgbcodec}).
Secondly, a sparse spatial propagation module that is likewise used within the encoder and decoder of the sparse-to-dense codec (\cref{sec:sparsespatialconv}).
Thirdly, the u-shaped sparse-to-dense network that applies the propagation module for guidance and considers sparsity throughout (\cref{sec:sparsecodec}).
Lastly, a dense refinement module to further improve the dense result.
The combination of all elements -- our sparse-to-dense interpolation network -- is visualized in \cref{fig:network}.

\subsection{RGB Codec} \label{sec:rgbcodec}
The purpose of the RGB codec is to provide a well-shaped feature representation of the image that fits the according level of the sparse codec.
Therefore it mimics the shape of the sparse codec and has the same number of levels $l$ in the encoder and decoder as the interpolator.
The image gets pre-processed by a regular $1\times1$ convolution and is then passed through $l$ down-sampling blocks.
Each consists of four $3\times3$ convolutions where the third convolution applies a stride of 2 to sub-sample the representation. 
After one additional convolution at the bottleneck, the representation of lowest resolution is passed through $l$ up-sampling blocks.
Again, each of these blocks consists of four $3\times3$ convolutions, but this time the second one is a transposed convolution with a stride of 2 for up-sampling.
In addition after up-sampling, the intermediate feature representation gets concatenated with the next higher resolved level of the encoder, \ie regular skip connections to re-introduce localization into the feature maps.
In this architecture, the number of output channels is gradually increased as the spatial resolution is reduced which is a common practice for low resolution feature embeddings.
In our setup, we use $l=6$ pyramid levels with fully symmetric feature depth of 32, 32, 48, 64, 80, 96, and 128.
An overview of the RGB codec is shown in \cref{fig:network:overview}.

Finally, we branch two affinity blocks from each level of the decoder to predict the spatially-variant, content-dependent kernels for each scale.
One affinity block consists of two convolutional layers. One layer is used for pre-transformation, and one to predict a single $K \times K$ kernel per pixel for propagation in the sparse-to-dense codec.
Please note, that different sets of propagation kernels are predicted for the encoder and the decoder of the sparse codec, \ie weights are not shared for the two affinity blocks at each level of the RGB decoder.
For reasons of memory consumption and computational efficiency, our propagation kernels have a size of $K=3$.
Contrary to existing work \cite{cheng2018depth}, our network uses a single, flat affinity map independent of the number of feature channels to propagate.
This reduces the total number of parameters significantly and effectively diminishes over-fitting during fine-tuning on small data sets.

\subsection{Sparse Spatial Propagation} \label{sec:sparsespatialconv}
The previously computed multi-scale feature maps, affinity maps, and propagation kernels are now used within our sparse spatial propagation module.
Consider an arbitrarily shaped $H \times W \times C$ feature representation $\mathcal{S}$ of the sparse input along with a binary sparsity mask $\mathcal{M}$ of shape $H \times W \times 1$ and a feature representation $\mathcal{F}$ of the guidance image of the same spatial size (and potentially a different number of feature channels).
The affinity block of the previous section will transform the image features $\mathcal{F}$ into a set of propagation kernels $\mathcal{K}$ of the shape $H \times W \times 1 \times K^2$.
For the sake of affinity and propagation, the center pixel of the propagation kernels is fixed to 1, \ie isolated sparse points will not be altered.
These kernels are then applied in a channel-wise $K \times K$ convolution with the sparse representation $\mathcal{S}$ to spread the information into the neighborhood according to the image features.
In GuideNet \cite{tang2019learning} one set of kernels is predicted for each feature channel of the sparse input, which leads to the necessity of depth-wise separable convolutions \cite{chollet2017xception}.
Other than that, we predict a single affinity map, which results in the natural use of depth-wise convolution for practicability and efficiency.
After channel-wise spatial propagation, a $1 \times 1$ convolution is performed to mix the propagated input dimension and expand (or compress) the representation to a new feature depth.
Further and in contrast to existing methods using convolutional spatial propagation, we explicitly model sparsity-awareness in our propagation module.
Towards this end, we adopt the idea of sparse convolution from \cite{uhrig2017sparsity} and utilize the sparsity mask $\mathcal{M}$ to normalize the propagated features.
By that, only valid information is spread according to the guidance image to fill in gaps.
Formally, the output of the sparse spatial convolution of $\mathcal{S}$ with $\mathcal{K}$ for a single channel $c$ and pixel is
\begin{equation}
	\tilde{\mathcal{S}}_c = \frac{\sum_{i,j \in \mathcal{W}} \mathcal{S}_{c,i,j} \cdot \mathcal{K}_{i,j}}{\sum_{i,j \in \mathcal{W}} \mathcal{M}_{i,j}},
\end{equation}
where $\mathcal{W}$ is the $k \times k$ window around the pixel under consideration.
The normalization and the propagation kernel are independent of the feature channel, \ie there are only a single 1-channel mask $\mathcal{M}$ and single set of kernels $\mathcal{K}$ for the entire feature volume.
This relationship is also visualized in \cref{fig:network:sparsespatialpropagation}.
The entire concept expands directly to arbitrary batch sizes.

\subsection{Image-guided Sparse-to-Dense Codec} \label{sec:sparsecodec}
The RGB codec and the sparse spatial propagation module enable an efficient way to introduce image guidance to our interpolation network.
All convolutions of the sparse-to-dense codec make use of the sparse convolution as presented by \cite{uhrig2017sparsity}.
Sparsity masks are used throughout the entire sparse codec which makes it easy to verify that full density is reached by the end of the decoder by the latest (usually already at the bottleneck), \ie all pixels have been filled with information from the initially valid points.
As with the RGB codec, we pre-process the sparse input with a sparse $1 \times 1$ convolution.
Then, $l$ sparse down-sampling blocks are applied.
These blocks consist of our sparse spatial propagation module that applies the spatial guidance kernels from the RGB decoder, followed by a $1 \times 1$ convolution to complete the depth-wise separation of the spatially variant guidance.
The last step within this block is a sparse average pooling layer with a kernel size of $3 \times 3$ and a stride of 2 to perform the sparse sub-sampling.
Again, a single $3 \times 3$ convolution is applied at the bottleneck.
Starting at lowest resolution from the bottleneck, $l$ guided up-sampling blocks are passed through. As with the down-sampling, the first part of these blocks is the depth-wise separated sparse spatial propagation.
Then, the feature representation along with its validity mask are up-sampled using nearest-neighbor interpolation to avoid mixture with invalid pixels in case some are still remaining.
Lastly, skip connections are established from the next higher resolution of the sparse encoder.
The skipped encoder features are summed up with the decoder features to avoid re-introduction of sparsity into the feature representation and merged in another $3 \times 3$ convolution.

At full input resolution of the decoder pyramid, we perform one additional sparse spatial guided propagation, followed by three more convolutions for final decoding.
The first two of these three are of size $3 \times 3$, the other is $1 \times 1$. The last two have linear activation to allow a final prediction of negative motions.
We are aware that, theoretically, the two linear activated convolutions could be folded into a single one.
However, we found that explicit separation leads to a faster convergence initially, probably due to better initialization by separation.
Another advantage of using sparse convolution is that (especially during the decoding) no negative boundary effects are introduced, because the sparsity mechanism can treat padded areas as invalid.

\subsection{Dense Refinement} \label{sec:refinement}
At the end of the sparse-to-dense codec, a dense result in the respective target domain is already obtained.
However, we follow the idea of CSPN \cite{cheng2018depth} and further refine the result using spatial propagation for filtering.
Since the RGB codec provides already a strong feature representation, we can transform these features into affinity maps for each output channel using a single $3 \times 3$ convolution.
The kernels extracted from the affinity maps are further transformed to introduce stability as in CSPN \cite{cheng2018depth}.
The dense results are then refined during 10 iterations of spatial propagation.

\subsection{Data, Training, and Implementation Details} \label{sec:data}
\paragraph{Data Sets.}
For real applications, realistic data is required.
However, labeling real world data with reference displacement fields is non-trivial and sometimes even impossible.
Therefore, only a limited amount of suitable data sets is available.
Additionally, these data sets are small in size, \ie in the number of distinct images.
This work will mainly use the KITTI 2015 data set \cite{menze2015object} to cover realistic scenarios which only provides 200 annotated images for scene flow and optical flow.
To overcome this issue, we will make use of synthetically generated data, namely the FlyingThings3D (FT3D) data set \cite{mayer2016large}.
It provides approximately $2500$ sequences, with 10 images each, of 3D objects flying in front of a random background image.
This data set is large enough for deep training, but lacks variation in the scenes and realism. Still, it has been shown to be irreplaceable for pre-training \cite{ilg2017flownet,mayer2016large,saxena2019pwoc,sun2018pwc}.
Next to KITTI and FT3D, Sintel \cite{butler2012sintel} provides a trade-off between realism and size, though only for optical flow.
Sintel comprises 23 sequences of 20 to 50 frames each.
Additionally, we use HD1K \cite{kondermann2016hci} for extended experiments with interpolation of optical flow.
For depth completion, the KITTI Benchmark Suite \cite{geiger2012kitti,uhrig2017sparsity} offers a larger and yet more realistic data set that provides labels for about $45000$ stereo image pairs.

For all results in \cref{sec:results}, we follow the common recommendation and perform our experiments on a randomly selected validation split which is not used for training.
In particular these sets are the 20 sequences 4, 42, 46, 65, 92, 94, 98, 106, 115, 119, 121, 124, 146, 173, 174, 181, 184, 186, 190, 193 on KITTI, the original \textit{val\_selection\_cropped} split from the KITTI depth completion data, the sequences \textit{alley\_2}, \textit{ambush\_4}, \textit{bamboo\_2}, \textit{cave\_4}, \textit{market\_5} for Sintel that sum up to 223 frames, and the sequences 0, 5, 15, 16, 18, 19, 27, 31 for HD1K.

\paragraph{Details.}


For large size data sets like FT3D, it is infeasible to compute the actual sparse input of existing sparse-to-dense pipelines, due to the high run-times of several seconds up to one minute per frame.
Instead and because FT3D is only used for pre-training, a randomized sparsification process is introduced to simulate the sparse or non-dense input for interpolation.
%
%
Additionally, random Gaussian noise ($\sigma=2$ px) is added to all remaining valid pixels to simulate inaccuracies of a real matching process.
For our experiments on optical flow and scene flow interpolation, we first train our network on FT3D \cite{mayer2016large}. The KITTI depth completion data set is sufficiently large to train on it directly.
We pre-train for 1 million iterations which corresponds to approximately 64 epochs.
Afterwards, we start training on the respective target domain and task with the pre-trained weights for initialization.
For pre-training, photometric image augmentation is applied as in \cite{dosovitskiy2015flownet}.
The objective for training depends on the specific interpolation problem at hand.
For motion fields, the average Euclidean distance between predicted $\mathbf{\hat{p}}$ and ground truth $\mathbf{p}$  motion vectors is minimized.
This loss function is equally used for optical flow and scene flow.
For single valued depth, we optimize the mean squared error between ground truth $d$ and prediction $\hat{d}$.
Except for the two final linearly activated layers, we use ReLU activation \cite{glorot2011deep} for all convolutional layers.
ADAM \cite{kingma2015adam} with an initial learning rate of $10^{-4}$ is used. The learning rate is continuously reduced with an exponential decay rate of $0.8$ after every 10 \% of the total number of steps.
Due to hardware constraints, we are limited to a batch size of 1 for all our experiments.
For training stability and improved generalization, we normalize all input of our network  according to the respective image and sparse statistics to zero mean and unit variance.

\section{Experiments and Results} \label{sec:results}
Three sets of experiments are presented.
The first one is an ablation study on the different components of the architecture to clarify our contributions and validate the impact.
Then, we demonstrate the robustness of \name{} in terms of noisy input, wrong input, changes of density of the input, and padding artifacts.
Lastly, \name{} is compared to state-of-the-art on various data sets and interpolation tasks.

For flow interpolation, the metrics under considerations are the end-point error (EPE) in image space, and the KITTI outlier error rate (KOE) giving the percentage of pixels that exceed an EPE of 3 px and deviate more than 5~\% from the ground truth. Both metrics are likewise applied in our experiments on scene flow and optical flow.
For depth completion, we use the default mean absolute error (MAE) and the root mean squared error (RMSE) as measure.

To obtain the sparse input for our experiments with optical flow, we use the prominent FlowFields (FF) \cite{bailer2015flow} or its extension FlowFields+ (FF+) \cite{bailer2019flow} along with their competitor CPM \cite{hu2016efficient}.
There has also been a longer history of sparse matching techniques in optical flow \cite{he2012computing,weinzaepfel2013deepflow}. However latest interpolation approaches \cite{hu2017robust,revaud2015epic} have shown that these have been superseded by the FlowFields family or CPM. 
Their matching concept has been extended to a stereo camera setup to predict scene flow correspondences in SceneFlowFields (SFF) \cite{schuster2018sceneflowfields} and further to a multi-frame setup in SceneFlowFields++ (SFF++) \cite{schuster2020sffpp}.
To the best of our knowledge, these are the only approaches which have tested the sparse-to-dense approach for scene flow.
For the problem of depth completion, sparse input is obtained directly from a LiDAR sensor.

\subsection{Ablation Study} \label{sec:results:ablation}
Part of our contributions is the combination of sparsity-awareness and spatial propagation for full guidance into an end-to-end interpolation network.
Therefore, in this section our approach is compared to equivalent networks that differ only conceptually from our design.
All the results of the ablation study are reported in \cref{tab:ablation}.
As a first step, we will validate that the fusion of image data into the sparse target domain (image guidance) is beneficial, especially when image data is available anyways.
Towards that goal, we evaluate an \textit{unguided} version of the sparse-to-dense codec, \ie the input image is not used at all and the RGB branch is removed.
Whenever the ablation removes our Sparse Spatial Guided Propagation, we replace it with a spatially invariant $3 \times 3$ convolution.
We also test different variants of guidance.
We remove guidance from either the encoder or decoder of the sparse-to-dense codec and compare to our fully guided approach.
It is obvious that guidance improves the results significantly.
Furthermore, guidance in the encoder alone (\textit{enc}) performs not as good as in later stages of the network (\textit{dec}), or during all stages (\textit{full}).
The latter two variants perform on a par, but we argue that full guidance improves results in difficult scenarios without much additional computational effort.

Next, we compare networks that use regular convolution wherever our design uses sparse convolution (\textit{sparse}) and networks which compute either a full affinity volume for guidance or a single affinity map (\textit{flat}).
Because LiDAR measurements have a quite regular pattern across all samples, the network variants without sparse convolution perform in general slightly better than our versions with sparse convolution.
Anyways, we will show in \cref{sec:results:robustness} that sparse convolution introduces higher robustness in case this property is not fulfilled.
The flat versions reduce the network size and computational complexity by more than 50~\% without much loss of accuracy.
In fact, the version with flat guidance and regular convolutions performs the best.
In later experiments with smaller data sets, we found the impact of flat guidance to be even more beneficial to reduce over-fitting.
Lastly, we show that dense \textit{refinement} improves the results for all variants with very little increase in number of parameters or FLOPs.

The fifth row in \cref{tab:ablation} represents a setup which is conceptually comparable to GuideNet \cite{tang2019learning}, \ie guidance is only used in the encoder, the network is not sparsity-aware, and guided propagation uses the full affinity volume. We call this setup \textit{GuideNet-like}.

\begin{table}[t]
	\centering
	\caption{Ablation study. We compare different concepts for sparse-to-dense interpolation of LiDAR measurements on the validation split of KITTI data. Mean absolute error (MAE) [mm], root mean squared error (RMSE) [mm], number of parameters ($\times10^6$) and floating point operations ($\times10^9$) are presented.}
	\label{tab:ablation}
	\vspace{1.5mm}
	\resizebox{1\linewidth}{!}{
	\begin{tabular}{cccc|cccc}
		 Guide & Sparse & Flat & Refine & MAE & RMSE & Params & FLOPs\Bstrut\\
		\hline
		\textbf{none} & yes & yes & no & 356 & 1171 & \textbf{0.93} & \textbf{41.2}\Tstrut\\ 	
		\textbf{enc} & yes & yes & no & 312 & 1013 & 4.32 & 148.5\\ 			
		\textbf{dec} & yes & yes & no & 289 & \textbf{953} & 4.47 & 149.5\\			
		\textbf{full} & yes & yes & no & \textbf{288} & 957 & 4.61 & 156.9\Bstrut\\	
		\hline
		enc & no & no & no & 280 & 929 & 6.49 & 250.1\Tstrut\Bstrut\\
		\hline
		full & \textbf{yes} & \textbf{no} & no & 276 & 915 & 10.14 & 382.4\Tstrut\\		
    	full & \textbf{no} & \textbf{no} & no & 270 & 910 & 10.14 & 381.3\\	
		full & \textbf{yes} & \textbf{yes} & no & 288 & 957 & \textbf{4.61} & 156.9\\	
		full & \textbf{no} & \textbf{yes} & no & \textbf{267} & \textbf{908} & \textbf{4.61} & \textbf{155.8}\Bstrut\\	
		\hline
    	full & yes & no & \textbf{yes} & 260 & 892 & 10.15 & 384.7\Tstrut\\
    	full & no & no & \textbf{yes} & 251 & 881 & 10.15 & 383.6\\
		full & yes & yes & \textbf{yes} & 260 & 910 & \textbf{4.61} & 159.2\\
    	full & no & yes & \textbf{yes} & \textbf{248} & \textbf{877} & \textbf{4.61} & \textbf{158.1}\\
	\end{tabular}
	}
\end{table}

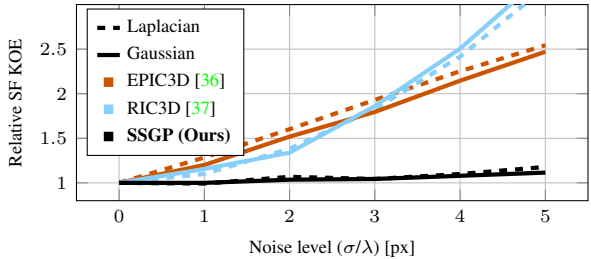
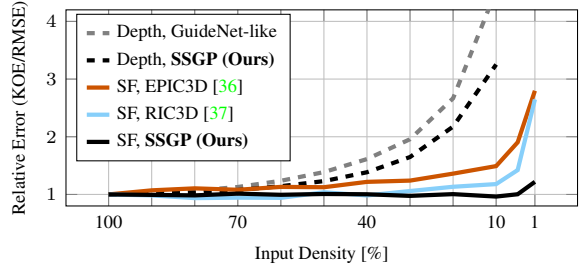
\begin{figure*}
	\centering
	\begin{subfigure}[t]{0.48\linewidth}
		\centering
		\begin{tikzpicture}
			\tikzstyle{every node}=[font=\scriptsize]
			\begin{axis}[
		        width=\linewidth,
		        height=0.5\linewidth,
				xtick={0,1,2,3,4,5},
    			ymin = 0.8, ymax = 3,
				ytick={1,1.5,2,2.5},
				grid=both,
		        xlabel={Noise level ($\sigma$/$\lambda$) [px]},
		        ylabel={Relative SF KOE},
    			ylabel near ticks,
    			xlabel near ticks,
				legend style={at={(0.02,0.98)}, anchor=north west}, 
				legend cell align={left},
  				legend image post style={scale=0.5},
			    every axis plot/.append style={ultra thick},
				legend entries={Laplacian\\
                				Gaussian\\
                				EPIC3D \cite{schuster2018sceneflowfields}\\
                				RIC3D \cite{schuster2020sffpp}\\
                				\textbf{SSGP (Ours)}\\}
           	]
			\addlegendimage{dashed, black}
			\addlegendimage{solid, black}
			\addlegendimage{only marks,mark=square*,color=burntorange}
			\addlegendimage{only marks,mark=square*,color=skyblue}
			\addlegendimage{only marks,mark=square*,color=black}
		        
		    \addplot[dashed,color=burntorange] plot coordinates {
		        (0, 27.4/27.4)
		        (1, 35.2/27.4)
		        (2, 43.9/27.4)
		        (3, 52.9/27.4)
		        (4, 61.7/27.4)
		        (5, 69.7/27.4)
		    };
		
		    \addplot[dashed,color=skyblue] plot coordinates {
		        (0, 18.0/18.0)		    	
		        (1, 19.8/18.0)
		        (2, 24.9/18.0)
		        (3, 32.9/18.0)
		        (4, 43.5/18.0)
		        (5, 57.5/18.0)
	        };
		
		    \addplot[dashed,color=black] plot coordinates {
		        (0, 25.2/25.2)
		        (1, 25.0/25.2)
		        (2, 26.9/25.2)
		        (3, 26.3/25.2)
		        (4, 27.7/25.2)
		        (5, 29.7/25.2)
	        };
		        
		    \addplot[color=burntorange] plot coordinates {
		        (0, 27.4/27.4)
		        (1, 32.9/27.4)
		        (2, 41.6/27.4)
		        (3, 49.2/27.4)
		        (4, 58.8/27.4)
		        (5, 67.7/27.4)
		    };
		
		    \addplot[color=skyblue] plot coordinates {
		        (0, 18.0/18.0)
		        (1, 20.8/18.0)
		        (2, 24.1/18.0)
		        (3, 33.5/18.0)
		        (4, 45.1/18.0)
		        (5, 60.2/18.0)
	        };
		
		    \addplot[color=black] plot coordinates {
		        (0, 25.2/25.2)
		        (1, 25.2/25.2)
		        (2, 26.1/25.2)
		        (3, 26.3/25.2)
		        (4, 27.2/25.2)
		        (5, 28.1/25.2)
	        };
		    
			\end{axis}
		\end{tikzpicture}
		\caption{Results for scene flow interpolation when the input is superposed with different types and levels of random noise.}
		\label{fig:robustness:noise}
	\end{subfigure}%
	\hfill%
	\begin{subfigure}[t]{0.48\linewidth}
		\centering
		\begin{tikzpicture}
			\tikzstyle{every node}=[font=\scriptsize]
			\begin{axis}[
		        width=\linewidth,
		        height=0.5\linewidth,
				xtick={1.0, 0.9, 0.8, 0.7, 0.6, 0.5, 0.4, 0.3, 0.2, 0.1, 0.01},
				xticklabels = {100, , , 70, , , 40, , , 10, 1},
				x dir=reverse,
    			ymin = 0.8, ymax = 4.2,
				ytick={1.0,2.0,3.0,4.0,5.0},
				grid=both,
		        xlabel={Input Density [\%]},
		        ylabel={Relative Error (KOE/RMSE)},
    			ylabel near ticks,
    			xlabel near ticks,
				legend style={at={(0.02,0.98)}, anchor=north west}, 
				legend cell align={left},
  				legend image post style={scale=0.5},
			    every axis plot/.append style={ultra thick},
				legend entries={Depth, GuideNet-like\\
                				Depth, \textbf{SSGP (Ours)}\\
                				SF, EPIC3D \cite{schuster2018sceneflowfields}\\
                				SF, RIC3D \cite{schuster2020sffpp}\\
                				SF, \textbf{SSGP (Ours)}\\}
            ]
			\addlegendimage{dashed, color=gray}
			\addlegendimage{dashed, color=black}
			\addlegendimage{solid, burntorange}
			\addlegendimage{solid, skyblue}
			\addlegendimage{solid, black}
		 
%
%
%
	
		    \addplot[dashed,color=gray] plot coordinates {
				(1.0, 0.93/0.93)
				(0.9, 0.95/0.93)
				(0.8, 0.99/0.93)
				(0.7, 1.05/0.93)
				(0.6, 1.15/0.93)
				(0.5, 1.29/0.93)
				(0.4, 1.50/0.93)
				(0.3, 1.82/0.93)
				(0.2, 2.48/0.93)
				(0.1, 4.22/0.93)
	        };
		           
		    \addplot[dashed,color=black] plot coordinates {
				(1.0, 0.91/0.91)
				(0.9, 0.93/0.91)
				(0.8, 0.94/0.91)
				(0.7, 0.98/0.91)
				(0.6, 1.04/0.91)
				(0.5, 1.12/0.91)
				(0.4, 1.26/0.91)
				(0.3, 1.50/0.91)
				(0.2, 1.98/0.91)
				(0.1, 2.96/0.91)

		    };
		   	
		   	\addplot[solid,color=burntorange] plot coordinates {
				(1.0, 27.4/27.4)
				(0.9, 29.34/27.4)
				(0.8, 30.29/27.4)
				(0.7, 29.61/27.4)
				(0.6, 30.98/27.4)
				(0.5, 30.82/27.4)
				(0.4, 33.37/27.4)
				(0.3, 33.97/27.4)
				(0.2, 37.28/27.4)
				(0.1, 40.86/27.4)
				(0.05, 52.16/27.4)
				(0.01, 76.65/27.4)

		    };
		    
		    \addplot[solid,color=skyblue] plot coordinates {
				(1.0, 18.0/18.0)
				(0.9, 17.7/18.0)
				(0.8, 16.85/18.0)
				(0.7, 16.99/18.0)
				(0.6, 16.94/18.0)
				(0.5, 18.67/18.0)
				(0.4, 17.69/18.0)
				(0.3, 19.06/18.0)
				(0.2, 20.39/18.0)
				(0.1, 21.23/18.0)
				(0.05, 25.6/18.0)
				(0.01, 47.65/18.0)

		    };
		    
		    \addplot[solid,color=black] plot coordinates {
				(1.0, 25.2/25.2)
				(0.9, 25.0/25.2)
				(0.8, 24.79/25.2)
				(0.7, 25.55/25.2)
				(0.6, 25.04/25.2)
				(0.5, 25.4/25.2)
				(0.4, 25.27/25.2)
				(0.3, 24.57/25.2)
				(0.2, 25.33/25.2)
				(0.1, 24.27/25.2)
				(0.05, 25.28/25.2)
				(0.01, 30.71/25.2)

		    };
		    		    
			\end{axis}
		\end{tikzpicture}
		\caption{Relative increase of errors when the input for different interpolation tasks is uniformly sparsified.}
		\label{fig:robustness:density}
	\end{subfigure}
	\vspace{1.5mm}
	\caption{Experiments on the robustness of \name{}. We alter the input with additive Gaussian and Laplacian noise (\subref{fig:robustness:noise}) or random sparsification for depth completion and interpolation of scene flow (\subref{fig:robustness:density}). Our novel architecture is most robust to any type or level of degradation.}
	\label{fig:robustness}	
\end{figure*}

\subsection{Robustness} \label{sec:results:robustness}
In this section, the robustness of \name{} is demonstrated.
We evaluate \name{} when the input is deteriorated with random noise, and when the density is reduced by random sampling.
Both results are presented in \cref{fig:robustness}.
For the experiment with noisy input, we add random Gaussian or Laplacian noise with zero mean and different values of standard deviation $\sigma$ and exponential decay $\lambda$ to all valid points of the sparse input.
We then perform scene flow interpolation and compare the relative increase of outliers for different levels of noise and different interpolation approaches with respect to the unaltered input.
\Cref{fig:robustness:noise} clearly shows, that our \name{} is extremely robust even to very noisy input. The outlier rate is maintained almost constant, while the competing methods perform considerably worse even for small amounts of additive noise.

In a second experiment, we also validate that the contribution of sparse convolution during guided propagation and the rest of the sparse-to-dense codec introduces higher invariance to the level of sparsity.
Towards this end, we perform depth completion and scene flow interpolation with randomly sparsified input. Results are presented in \cref{fig:robustness:density}.
The increase of errors for the sparsity-aware model is about 50~\% less when considering very sparse depth measurements. For \name{} on scene flow (\textit{SF}), the impact of sparisfication is neglectable until 1~\% of the original density.
Note that all models are trained on the full input density.
This improved robustness applies also to changes in the pattern of the input, \eg when the LiDAR measurements are sparsified non-uniformly.

As additional indicator for the robustness of \name{}, we measure the \textit{outlier rejection rate} (ORR), \ie the percentage of input that is classified as scene flow outlier before interpolation, but is corrected during interpolation.
For input from SFF and SFF++, EPIC3D achieves ORRs of 51.2 \% and 40.3 \%, RIC3D achieves 64.2 \% and 55.7 \%, and our \name{} yields ORRs of \textbf{67.6 \%} and \textbf{56.7 \%}.

We also compare the errors at boundary regions of the image to show the robustness of sparse convolution to padding. While the \textit{GuideNet-like} variant obtains an MAE and RMSE of 186 and 505 mm in regions which are less than 10 px away from the image boundary, our full setup of \name{} achieves \textbf{140} and \textbf{448 mm}.

\begin{table}
	\centering
	\caption{Evaluation of \textit{scene flow} interpolation on our validation split of the KITTI scene flow data set. KITTI outliers (KOE) [\%], end-point error (EPE) [px], and run time [s] are reported.}
	\label{tab:sf}
	\resizebox{1\linewidth}{!}{
	\begin{tabular}{cc||cc|cc|cc|cc|c}
		  & & \multicolumn{2}{c|}{D0} & \multicolumn{2}{c|}{D1} & \multicolumn{2}{c|}{OF} & \multicolumn{2}{c|}{SF} & Run\\  
		Input & Method & KOE & EPE & KOE & EPE & KOE & EPE & KOE & $\Sigma$EPE & time\Bstrut\\
		\hline
		\multirow{3}{*}{\rotatebox[origin=c]{90}{SFF}} & EPIC3D \cite{schuster2018sceneflowfields} & 12.83 & 1.88 & 17.80 & 11.49 & 29.62 & 112.1 & 31.72 & 125.4 & 1.0\Tstrut\\
		 & RIC3D \cite{schuster2020sffpp} & 9.88 & 1.92 & 13.94 & 2.79 & \textbf{15.44} & 8.42 & \textbf{17.45} & 13.10 & 3.8 \\
		 & \textbf{\name{} (Ours)} & \textbf{9.06} & \textbf{1.33} & \textbf{13.93} & \textbf{1.83} & 20.67 & \textbf{5.04} & 25.19 & \textbf{8.20} & \textbf{0.19}\Bstrut\\
		\hline
		\multirow{3}{*}{\rotatebox[origin=c]{90}{\begin{tabular}{c} SFF++\\ + SDC \end{tabular}}} & EPIC3D \cite{schuster2018sceneflowfields} & 6.74 & 1.30 & 10.83 & 1.96 & 15.65 & 6.23 & 17.91 & 9.49 & 1.0\Tstrut\\
		 & RIC3D \cite{schuster2020sffpp} & 5.91 & 1.29 & \textbf{7.24} & 1.53 & \textbf{9.80} & 3.33 & \textbf{11.50} & 6.15 & 3.8 \\
		 & \textbf{\name{} (Ours)} & \textbf{5.71} & \textbf{1.04} & 9.89 & \textbf{1.45} & 12.39 & \textbf{3.00} & 16.61 & \textbf{5.50} & \textbf{0.19} \\
	\end{tabular}
	}
\end{table}

\subsection{Interpolation} \label{sec:results:interpolation}

\paragraph{Scene Flow.}
As first application to our interpolation network, we use matches from SFF \cite{schuster2018sceneflowfields} and SFF++ \cite{schuster2020sffpp} (using the SDC feature descriptor \cite{schuster2019sdc}) for interpolation of dense scene flow.
The results are computed on the KITTI data set \cite{menze2015object} and are compared to EPIC3D \cite{schuster2018sceneflowfields} and RIC3D \cite{schuster2020sffpp} which are the heuristic two-stage interpolators of SFF and SFF++ respectively.
Both use additional edge information of the scene.
Results are given in \cref{tab:sf}.

Our approach achieves competitive performance to previous methods, though being significantly faster.
Especially for interpolation of initial disparity (\textit{D0}), \name{} outperforms the baselines.
Further, \name{} performs comparatively well in the EPE metric, which was also the objective function during training.

\begin{table*}[b]
	\caption{Comparison of methods for depth completion on the KITTI benchmark \cite{uhrig2017sparsity}. We report mean average error (MAE [mm]), root mean squared error (RMSE [mm]), and run time [ms] for the best performing, published methods using image guidance out of more than 90 total submissions. Values in gray are computed on the validation split.}
	\label{tab:depth}
	\hspace{0.01\linewidth}%
	\resizebox{0.93\linewidth}{!}{
	\begin{tabular}{r||C|C|C|C|C|C|C|C|C|C|C|C|C|C|C||C}
		 \multicolumn{1}{r}{ } & 
		 \rot{GuideNet \cite{tang2019learning}} & 
		 \rot{CSPN++ \cite{cheng2020cspn}} &
		 \rot{FuseNet \cite{chen2019learning}} &
		 \rot{DeepLiDAR \cite{qiu2019deeplidar}} &
		 \rot{MSG-CHN \cite{li2020multi}} &
		 \rot{Guide\&Certainty \cite{vangansbeke2019depth}} &  
		 \rot{PwP \cite{xu2019depth}} &
		 \rot{CrossGuidance \cite{lee2020deep}} &
		 \rot{Sparse-to-Dense \cite{ma2019self}} & 
		 \rot{NConv-CNN \cite{eldesokey2019confidence}} &
		 \rot{DDP \cite{yang2019dense}} &
		 \rot{\textbf{\name{} (Ours)}} & 
		 \rot{Spade \cite{jaritz2018sparse}} &
		 \rot{DFineNet \cite{zhang2019dfinenet}} & 
		 \rot{CSPN \cite{cheng2018depth}} & 
		 \rot{\textcolor{gray}{RIC3D \cite{schuster2020sffpp}}}\\
		 
		\textbf{MAE} & 219 & 209 & 221 & 227 & 220 & 215 & 235 & 254 & 250 & 233 & \textbf{204} & 245 & 235 & 304 & 279 & \textcolor{gray}{588}\\
		\textbf{RMSE} & \textbf{736} & 744 & 753 & 758 & 762 & 773 & 777 & 807 & 815 & 830 & 833 & 838 & 918 & 945 & 1020 & \textcolor{gray}{2477}\\
		\textbf{Run time} & 140 & 200 & 90 & 70 & \textbf{10} & 20 & 100 & 200 & 80 & 20 & 80 & 140 & 70 & 20 & 1000 & \textcolor{gray}{1400}\\
	\end{tabular}
	}
\end{table*}

\begin{figure}[t]
	\centering
	\hspace{0.08\linewidth}
	\begin{subfigure}[c]{0.45\linewidth}
		\adjincludegraphics[width=1\linewidth,trim={{.2\width} 0 {.3\width} {.3\height}},clip]{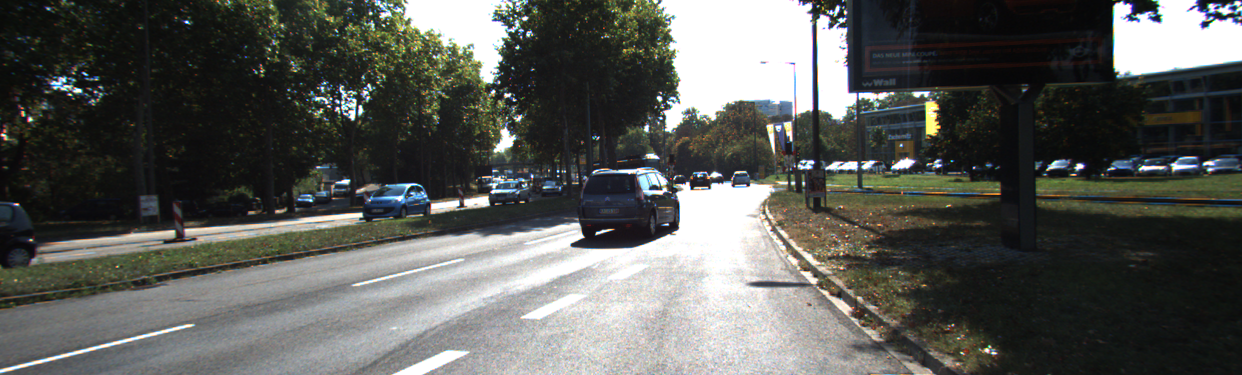}%
		\vspace{0.5mm}%
	\end{subfigure}
	\begin{subfigure}[c]{0.45\linewidth}
		\adjincludegraphics[width=1\linewidth,trim={{.2\width} 0 {.3\width} {.3\height}},clip]{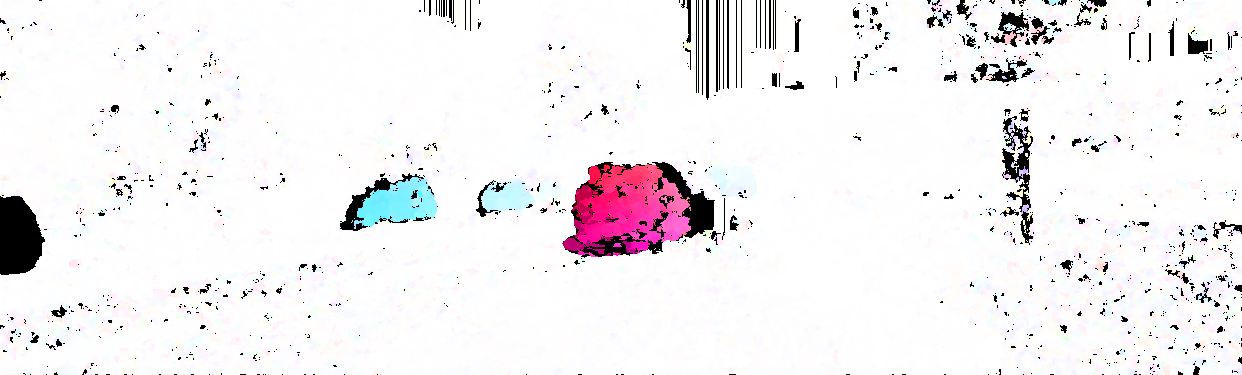}%
		\vspace{0.5mm}%
	\end{subfigure}\\
	\hspace{0.08\linewidth}
	\begin{subfigure}[c]{0.45\linewidth}
		\centering \footnotesize Image
	\end{subfigure}
	\begin{subfigure}[c]{0.45\linewidth}
		\centering \footnotesize FF+ \cite{bailer2019flow} Input
	\end{subfigure}\\
	\hspace{0.08\linewidth}
	\begin{subfigure}[c]{0.45\linewidth}
		\centering \footnotesize Dense Predictions
	\end{subfigure}
	\begin{subfigure}[c]{0.45\linewidth}
		\centering \footnotesize Error Maps
	\end{subfigure}\\
	\begin{subfigure}[c]{0.08\linewidth}
		\rotatebox[origin=c]{90}{\footnotesize \begin{tabular}{c} EPIC \\ \cite{revaud2015epic}\end{tabular}}
		\vspace{1mm}%
	\end{subfigure}
	\begin{subfigure}[c]{0.45\linewidth}
		\adjincludegraphics[width=1\linewidth,trim={{.2\width} 0 {.3\width} {.3\height}},clip]{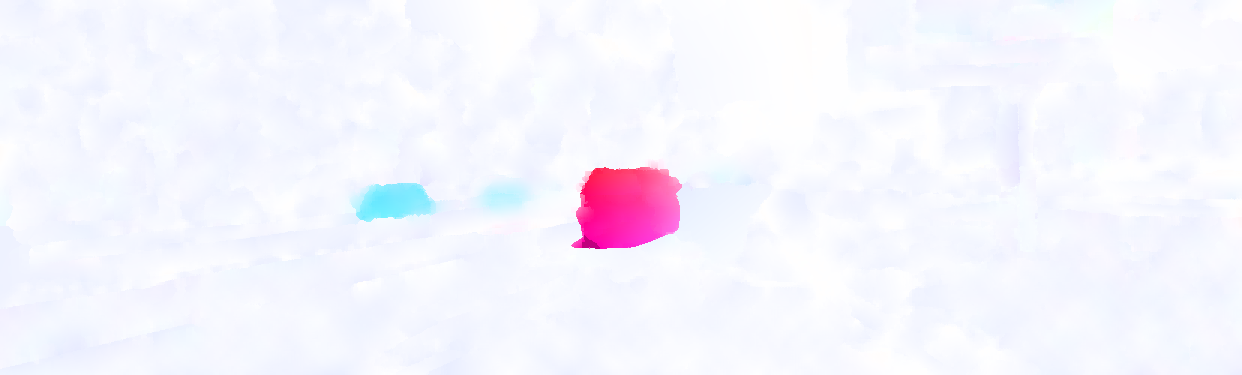}%
		\vspace{1mm}%
	\end{subfigure}
	\begin{subfigure}[c]{0.45\linewidth}
		\adjincludegraphics[width=1\linewidth,trim={{.2\width} 0 {.3\width} {.3\height}},clip]{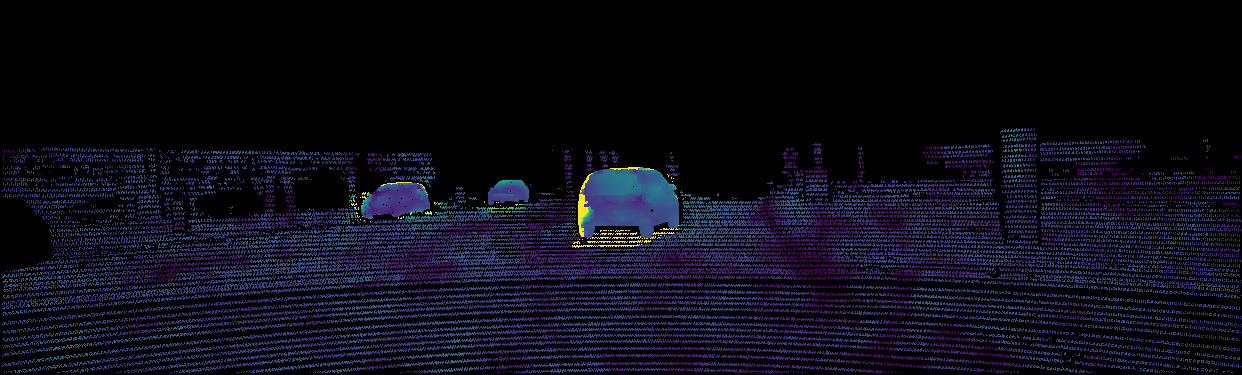}%
		\vspace{1mm}%
	\end{subfigure}\\
	\begin{subfigure}[c]{0.08\linewidth}
		\rotatebox[origin=c]{90}{\footnotesize \begin{tabular}{c} RIC \\ \cite{hu2017robust}\end{tabular}}
		\vspace{1mm}%
	\end{subfigure}
	\begin{subfigure}[c]{0.45\linewidth}
		\adjincludegraphics[width=1\linewidth,trim={{.2\width} 0 {.3\width} {.3\height}},clip]{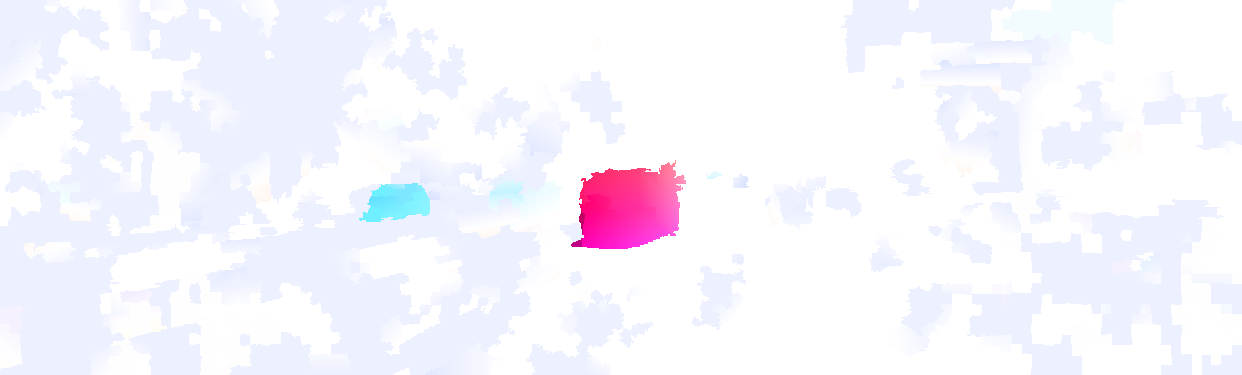}%
		\vspace{1mm}%
	\end{subfigure}
	\begin{subfigure}[c]{0.45\linewidth}
		\adjincludegraphics[width=1\linewidth,trim={{.2\width} 0 {.3\width} {.3\height}},clip]{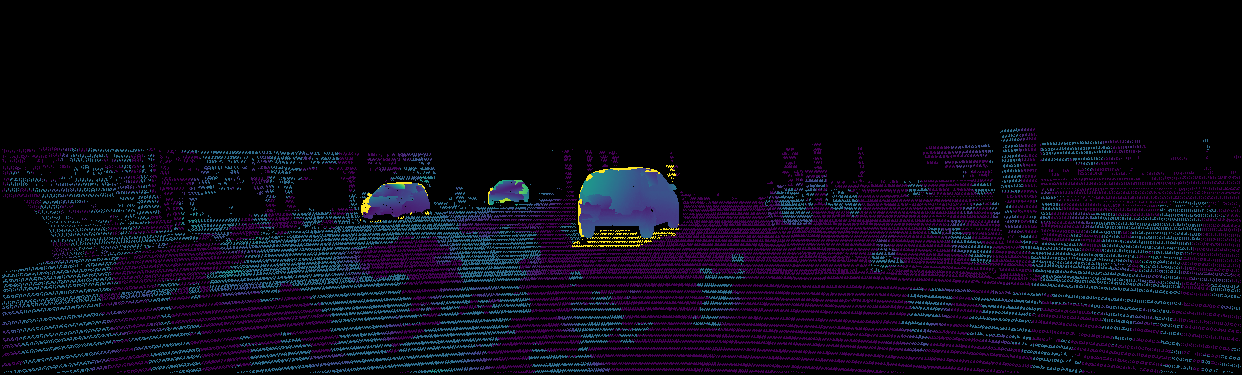}%
		\vspace{1mm}%
	\end{subfigure}\\
	\begin{subfigure}[c]{0.08\linewidth}
		\rotatebox[origin=c]{90}{\footnotesize \begin{tabular}{c} Interpo-\\ Net \cite{zweig2017interponet} \end{tabular}}
		\vspace{1mm}%
	\end{subfigure}
	\begin{subfigure}[c]{0.45\linewidth}
		\adjincludegraphics[width=1\linewidth,trim={{.2\width} 0 {.3\width} {.3\height}},clip]{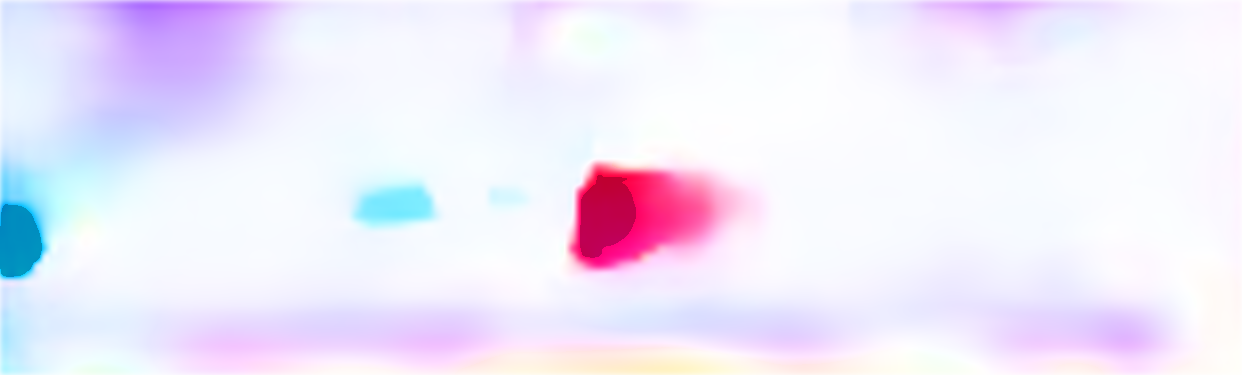}%
		\vspace{1mm}%
	\end{subfigure}
	\begin{subfigure}[c]{0.45\linewidth}
		\adjincludegraphics[width=1\linewidth,trim={{.2\width} 0 {.3\width} {.3\height}},clip]{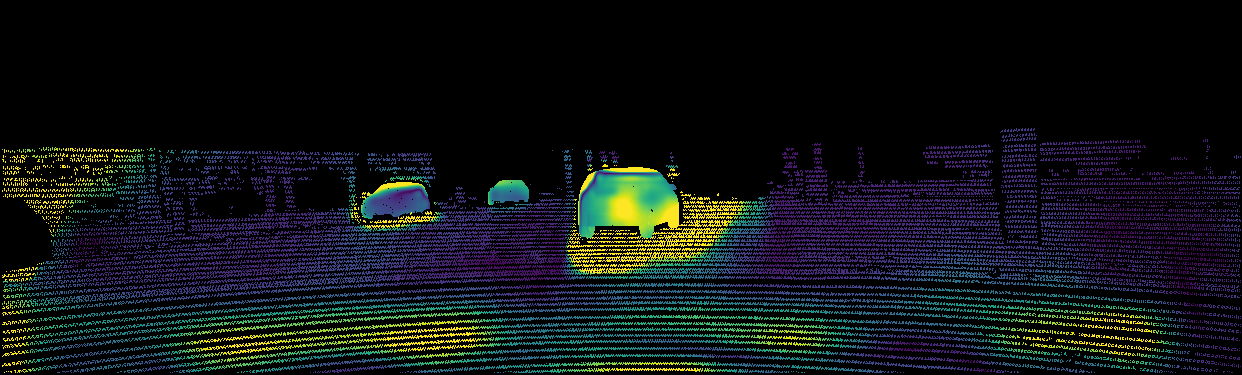}%
		\vspace{1mm}%
	\end{subfigure}\\
	\begin{subfigure}[c]{0.08\linewidth}
		\rotatebox[origin=c]{90}{\footnotesize \begin{tabular}{c} \textbf{\name{}} \\ \textbf{(Ours)} \end{tabular}}
	\end{subfigure}
	\begin{subfigure}[c]{0.45\linewidth}
		\adjincludegraphics[width=1\linewidth,trim={{.2\width} 0 {.3\width} {.3\height}},clip]{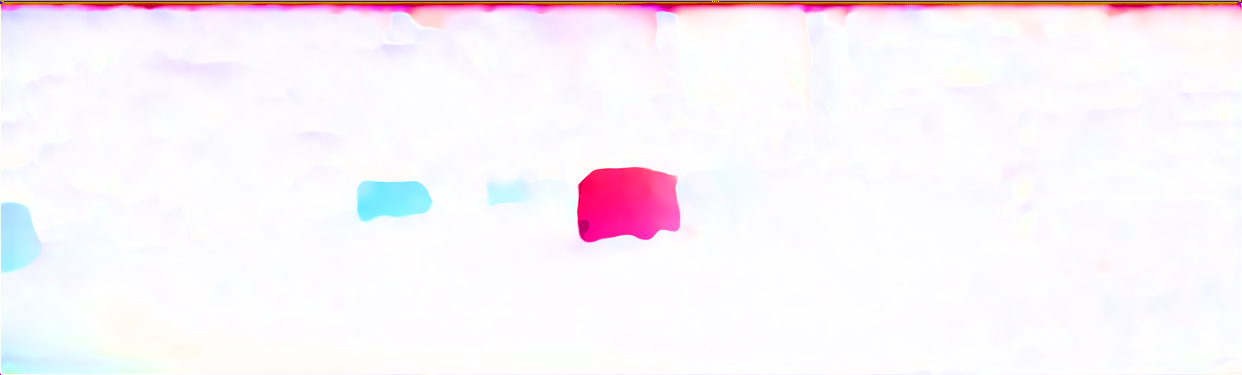}
	\end{subfigure}
	\begin{subfigure}[c]{0.45\linewidth}
		\adjincludegraphics[width=1\linewidth,trim={{.2\width} 0 {.3\width} {.3\height}},clip]{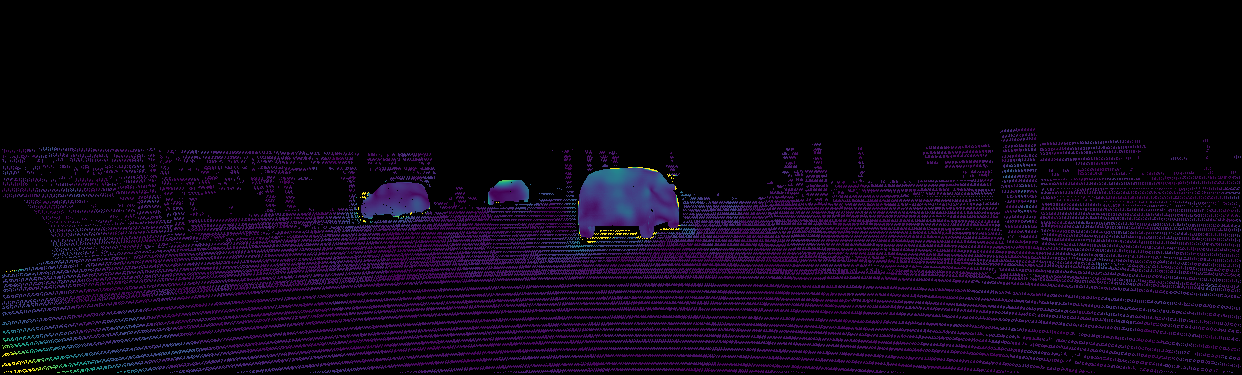}
	\end{subfigure}\\
	\begin{subfigure}[c]{0.1\linewidth}
		\footnotesize EPE:%
	\end{subfigure}
	\begin{subfigure}[c]{0.88\linewidth}
		\includegraphics[width=1\linewidth]{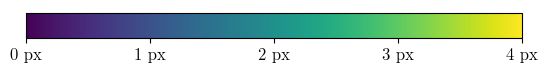}%
	\end{subfigure}\\
	\vspace{1mm}
	\caption{Visual comparison of optical flow interpolation on the KITTI data set.}
	\label{fig:results}
\end{figure}

\paragraph{Optical Flow.}
For the experiments related to optical flow, we have multiple data sets to evaluate on, namely KITTI \cite{menze2015object}, HD1K \cite{kondermann2016hci}, and Sintel \cite{butler2012sintel}.
We evaluate our method and state-of-the-art for two kinds of input matches generated from FF+ \cite{bailer2019flow} and CPM \cite{hu2016efficient}.
Our approach will be compared to EPICFlow \cite{revaud2015epic}, RICFlow \cite{hu2017robust}, and InterpoNet \cite{zweig2017interponet}.
Note, that all three methods use additional edge information, while we feed the raw image to our network.
A visual comparison for a cropped frame of KITTI is presented in \cref{fig:results}.
In this example, \name{} presents a globally consistent result, even in the static part of the scene, where small deviations have most impact in the visualization.
Our approach shows the most accurate and sharp object contours, even though it is not provided with pre-computed edge information.
This highlights the capabilities of the full guidance strategy.
In fact, our approach is able to reject wrong matches in shadows of the vehicles during interpolation.

\Cref{tab:of} compares quantitative results over our entire validation sets.
It is to highlight that \name{} cuts the end-point error on KITTI by about half in our comparison.
On KITTI also, the outlier rates of \name{} beat all previous work.
For completeness and fairness, we have to mention that we are using the publicly available pre-trained weights of InterpoNet \cite{zweig2017interponet} that have been fine-tuned on Sintel with input from DF \cite{menze2015discrete} and on KITTI with matches from FlowFields \cite{bailer2015flow}.
However, this indicates that InterpoNet is not very robust to changes of the input.
On Sintel, our approach is on par with InterpoNet, but lacks behind the other methods. This is due to the limited variance between scenes which makes it hard to train a deep model on Sintel.
Yet on HD1K, our \name{} outperforms state-of-the-art in all metrics while also being faster.


\paragraph{Depth Completion.}

\name{} can also be used for the completion of sparse \mbox{LiDAR} measurements.
We train the entire architecture from scratch on the KITTI depth completion data set \cite{uhrig2017sparsity} and compare our results to state-of-the-art in \cref{tab:depth}.
Our network again achieves a competitive result on yet another challenge, indicating its broad applicability.
A visual example of an interpolated depth map is given in \cref{fig:title}.
We further notice that RIC3D \cite{schuster2020sffpp}, a top-performing method for interpolation of scene flow, performs considerably worse than any other approach.
This shows, that even though RIC3D is not a learning-based method, it has a strong dependency on properly selected hyper-parameters.

\begin{table}[t]
	\centering
	\caption{Evaluation of interpolation of \textit{optical flow}. We test on our validation splits of the KITTI, HD1K, and Sintel data sets. Outlier rates (KOE) [\%], end-point error (EPE) [px], and run time [s] are reported.}
	\label{tab:of}
	\resizebox{1\linewidth}{!}{
	\begin{tabular}{cc||cc|cc|cccc|c}
	 	 & & & & & & \multicolumn{4}{c|}{Sintel} & \\
		 & & \multicolumn{2}{c|}{KITTI} & \multicolumn{2}{c|}{HD1K} & \multicolumn{2}{c}{clean} & \multicolumn{2}{c|}{final} & Run\\
		Input & Method & KOE & EPE & KOE & EPE & KOE & EPE & KOE & EPE & time\Bstrut\\
		\hline
		\multirow{4}{*}{\rotatebox[origin=c]{90}{CPM \cite{hu2016efficient}}} & EPICFlow \cite{revaud2015epic} & 24.39 & 10.04 & 5.43 & 1.11 & 9.98 & \textbf{3.84} & 13.94 & \textbf{5.76} & 0.4\Tstrut\\
		 & RICFlow \cite{hu2017robust} & 21.98 & 9.91 & 5.02 & 1.09 & \textbf{9.17} & 4.05 & \textbf{13.60} & 5.88 & 2.8 \\
		 & InterpoNet \cite{zweig2017interponet} & 40.38 & 12.81 & 12.3 & 2.36 & 14.94 & 4.75 & 18.09 & 6.24 & 0.3 \\
		 & \textbf{\name{} (Ours)} & \textbf{20.26} & \textbf{5.02} & \textbf{4.32} & \textbf{0.83} & 14.97 & 5.63 & 20.33 & 7.27 & \textbf{0.16} \Bstrut\\
		\hline
		\multirow{4}{*}{\rotatebox[origin=c]{90}{FF+ \cite{bailer2019flow}}} & EPICFlow \cite{revaud2015epic} & 23.97 & 11.34 & 5.55 & 1.21 & 11.25 & \textbf{5.05} & 15.99 & \textbf{7.26} & 0.4\Tstrut\\
		 & RICFlow \cite{hu2017robust} & 20.46 & 10.17 & 4.88 & 1.07 & \textbf{10.59} & 5.59 & \textbf{15.82} & 8.19 & 2.8 \\
		 & InterpoNet \cite{zweig2017interponet} & 37.08 & 11.34 & 13.1 & 2.35 & 16.49 & 5.7 & 20.51 & 7.64 & 0.3 \\
		 & \textbf{\name{} (Ours)} & \textbf{20.34} & \textbf{5.21} & \textbf{4.54} & \textbf{0.85} & 16.53 & 6.55 & 22.20 & 8.43 & \textbf{0.16} \\
	\end{tabular}
	}
\end{table}

\section{Conclusion} \label{sec:conclusion}
\name{} successfully combines sparsity-aware convolution and spatially variant propagation for fully image guided interpolation.
The network design is applicable to diverse sparse-to-dense problems and achieves competitive performance throughout all experiments, beating state-of-the-art in interpolation of optical flow and in terms of EPE.
A flat affinity map can be used for spatial guidance equally well as a full affinity volume, drastically reducing the overall network size.
This strategy for guidance resolves the dependency on explicitly pre-computed edge information resulting in even more accurate interpolation boundaries with a globally consistent output that preserves fine details.
\name{} is especially robust to variations of the sparsity pattern and to noise in the input.

\section*{Acknowledgement}
This work was partially funded by the BMW Group and partially by the Federal Ministry of Education and Research Germany under the project VIDETE (01IW18002).

{\small
\bibliographystyle{ieee_fullname}
\bibliography{bib}
}

\end{document}